\documentclass[review,12pt]{elsarticle}

\newcommand{\g}{\,\vert\,}

\newcommand{\indpt}{\protect\mathpalette{\protect\independenT}{\perp}}
\def\independenT#1#2{\mathrel{\rlap{$#1#2$}\mkern2mu{#1#2}}}
\newcommand{\E}[1]{\mathbb{E}\left[#1\right]}
\newcommand{\R}{\mathbb{R}}
\newcommand{\EE}[2]{\mathbb{E}_{#1}\left[#2\right]}

\newcommand{\cN}{\mathcal{N}}
\newcommand{\cD}{\mathcal{D}}
\newcommand{\rom}[1]{\uppercase\expandafter{\romannumeral #1\relax}}

\newcommand{\mbx}{\mathbf{x}}

\newcommand{\mbX}{\mathbf{X}}

\usepackage{amsmath}
\usepackage{caption}
\usepackage{subcaption}
\usepackage{hyperref}
\usepackage{xcolor}
\usepackage{booktabs}
\usepackage{multirow}
\usepackage{graphicx}

\newtheorem{assumption}{Assumption}

\newtheorem{definition}{Definition}
\newproof{proof}{Proof}

\usepackage[nameinlink]{cleveref}
\creflabelformat{equation}{#1#2#3}
\Crefname{equation}{Eq.}{Eqs.}
\Crefname{figure}{Fig.}{Figs.}
\Crefname{section}{Sec.}{Secs.}
\Crefname{assumption}{Assumption}{Assumptions}
\Crefname{appendix}{}{}

\usepackage{amssymb}

\journal{Journal of Biomedical Informatics}

\begin{document}

\begin{frontmatter}

\title{Causal Fairness Assessment of Treatment Allocation with Electronic Health Records}

\author[1]{Linying Zhang}
\author[1]{Lauren R.~Richter}
\author[2]{Yixin Wang}
\author[1]{Anna Ostropolets}
\author[1,4]{Noémie Elhadad}
\author[3,4]{David M.~Blei}
\author[1]{George Hripcsak}

\affiliation[1]{organization={Department of Biomedical Informatics, Columbia University Irving Medical Center},
            city={New York},
            state={NY},
            country={USA}}
\affiliation[2]{organization={Department of Statistics, University of Michigan},
            city={Ann Arbor},
            state={MI},
            country={USA}}      
\affiliation[3]{organization={Department of Statistics, Columbia University},
            city={New York},
            state={NY},
            country={USA}}
\affiliation[4]{organization={Department of Computer Science, Columbia University},
            city={New York},
            state={NY},
            country={USA}}

\begin{abstract}

\textbf{Objective}: Healthcare continues to grapple with the persistent issue of treatment disparities, sparking concerns regarding the equitable allocation of treatments in clinical practice. While various fairness metrics have emerged to assess fairness in decision-making processes, a growing focus has been on causality-based fairness concepts due to their capacity to mitigate confounding effects and reason about bias. However, the application of causal fairness notions in evaluating the fairness of clinical decision-making with electronic health record (EHR) data remains an understudied domain. This study aims to address the methodological gap in assessing causal fairness of treatment allocation with electronic health records data. In addition, we investigate the impact of social determinants of health on the assessment of causal fairness of treatment allocation.

\textbf{Methods}: We propose a causal fairness algorithm to assess fairness in clinical decision-making. Our algorithm accounts for the heterogeneity of patient populations and identifies potential unfairness in treatment allocation by conditioning on patients who have the same likelihood to benefit from the treatment. We apply this framework to a patient cohort with coronary artery disease derived from an EHR database to evaluate the fairness of treatment decisions. 

\textbf{Results}: Our analysis reveals notable disparities in coronary artery bypass grafting (CABG) allocation among different patient groups. Women were found to be 4.4\%-7.7\% less likely to receive CABG than men in two out of four treatment response strata. Similarly, Black or African American patients were 5.4\%-8.7\% less likely to receive CABG than others in three out of four response strata. These results were similar when social determinants of health (insurance and area deprivation index) were dropped from the algorithm. These findings highlight the presence of disparities in treatment allocation among similar patients, suggesting potential unfairness in the clinical decision-making process. 

\textbf{Conclusion}: This study introduces a novel approach for assessing the fairness of treatment allocation in healthcare. By incorporating responses to treatment into fairness framework, our method explores the potential of quantifying fairness from a causal perspective using EHR data. Our research advances the methodological development of fairness assessment in healthcare and highlight the importance of causality in determining treatment fairness.

\end{abstract}


\end{frontmatter}


\section{Introduction} \label{sec:intro}
Assessing fairness of treatment allocation is an important element of equitable health care. Decisions are made by clinicians on a daily basis that directly impact patient care, but how these decisions are made is a complex process. While the medical ideal is to base decisions on a patient's health condition, this is not the reality. Gender, race, ethnicity, socioeconomic status, and other sensitive attributes can influence clinicians' decision-making process, raising important concerns about inequity in health and health care \citep{dehon2017systematic, fitzgerald2017implicit, aberegg2004medical, vanryn2000effect}. 

Assessing fairness using observational data requires defining fairness quantitatively. There are many perspectives on how to quantify fairness. A major distinction among the existing fairness criteria is the use of causal inference. Associational fairness notions, such as statistical parity \citep{dwork2012fairness}, calibration \citep{chouldechova2017fair}, and accuracy, do not rely on causal reasoning and estimate fairness based on observed data alone. On the other hand, causal fairness notions, such as counterfactual fairness and path-specific causal fairness, rely on knowledge about the data generating process (e.g., a structural causal model) to assess fairness. Serious concerns have been raised about associational fairness because they ignore the confounding effect and as a result, multiple fairness notions cannot be simultaneously satisfied on a given dataset \citep{rahmattalabi2022promises, makhlouf2022survey, loftus2018causal}. Given the large number of existing fairness notions, what fairness metrics are appropriate for clinical settings remains a question. 

In medicine, the idea of fairness is expressed as \emph{health equity}. Health equity can be defined in multiple ways. According to \emph{Communities in Action: Pathways to Health Equity} \citep{nationalacademiesofsciencesengineeringandmedicine2017communities}, ``\textit{health equity is the state in which everyone has the opportunity to attain full health potential and no one is disadvantaged from achieving this potential because of social position or any other socially defined circumstance.}” This definition of health equity suggests the importance of considering the potential impact of different treatment options on patients' health when assessing health equity. In this paper, we adopt this definition of health equity.

Based on this definition of health equity, we explore the existing fairness notions to discover potential metrics for assessing the fairness of treatment allocation in healthcare. Our goal is to develop an algorithm for assessing treatment equity with electronic health records (EHRs). To illustrate our approach, we study the fairness in the allocation of coronary artery revascularization among patients diagnosed with coronary artery disease (CAD). 

Coronary artery disease is a major cause of mortality worldwide \citep{worldheartfederation2023world, tsao2023heart}. Percutaneous coronary intervention (PCI) and coronary artery bypass grafting (CABG) are commonly employed revascularization techniques for treating CAD. In most studies, CABG is more effective in comparison to PCI on multiple long-term health outcomes, including lower mortality rates and lower risk of myocardial infarction \citep{deb2013coronary, zhang2017percutaneous, serruys2009percutaneous, farkouh2012strategies}. However, CABG is much more expensive than PCI and requires longer hospital stay, which could pose a financial challenge to some patients. Despite recent guidelines that emphasize treatment decisions in patients with CAD should be based on clinical indication, regardless of sex, race, or ethnicity \citep{virani20232023, lawton20222021}, prior studies examining cardiovascular disparities have consistently identified that women and racial and ethnic minorities were less likely to undergo CABG, which has raised concerns regarding inequities in cardiovascular treatment \citep{angraal2018sex, heer2017sex, desai2019racial}. However, it is essential to acknowledge that these findings, while disconcerting, do not directly indicate unfairness in treatment allocation. This limitation arises from the fact that many of these studies primarily establish associations, often lacking comprehensive control over confounding factors such as disease severity. In this study, we develop a causal fairness algorithm to answer the question if there is unfairness in the allocation of CABG versus PCI in patients with coronary artery disease.

Our work contributes to the existing body of work on fairness in healthcare by bridging gaps between theoretical work on causal fairness and empirical research on treatment disparities from observational data. While previous research, as cited in our review of related works in the discussion section, has predominantly focused on associational fairness metrics, with some recent works focusing on causal fairness but are applied to the context of assessing fairness of risk prediction models, our work introduces a causal algorithm that retrospectively assesses the fairness of treatment decision-making using EHRs in clinical settings. Unlike \citet{qureshi2020causal} and \citet{pfohl2019counterfactual}, who primarily address fairness from a theoretical perspective or in the context of clinical risk prediction, our study delves into the practical application of these concepts, tailoring them to the unique challenges of treatment allocation in healthcare. Moreover, our algorithm, grounded in electronic health records, advances beyond the scope of studies like \citet{kusner2017counterfactual} and \citet{sun2022assessing} by incorporating a comprehensive set of patient characteristics, thereby addressing the intricate interplay between clinical and social determinants of health (SDoH) and treatment decision-making.

The contributions of this work are as follows. First, we develop a novel causal fairness algorithm for assessing the fairness of treatment allocation using electronic health records data with clearly articulated assumptions for identification. The algorithm is developed following the Observational Medical Outcomes Partnership (OMOP) common data model which enhances the algorithm's adaptability for investigating other diseases and populations. Second, we provide clinical insights on the fairness of treatment allocation in patients with coronary artery disease. Third, we demonstrate that social determinants of health (specifically, area deprivation index and insurance) have negligible impact on the causal fairness estimates when a large set of patient characteristics from EHRs are used in the fairness algorithm, shedding light on the intricate correlation between EHR features and social determinants of health.

\section{Methods}
\subsection{Definition of principal fairness}
Denote the variables in the treatment decision-making process as follows. For the $i$-th patient, let $A_i \in \{0,1\}$ be the sensitive attribute (e.g., gender, race), $D_i \in \{0,1\}$ be a  medical decision (e.g., CABG vs PCI), and $X_i \in \R^M$ be an $M$-dimensional vector of observed pre-treatment covariates (e.g. clinical conditions, medications, observations). Using the potential outcome framework, let $Y_i(d)$ be the potential value of the outcome corresponding to treatment $d$. There are two potential outcomes for each patient, $Y_i(0)$ and $Y_i(1)$. For example, $Y_i(0)=1$ means the $i$-th patient would have a heart attack within one year if not treated with heart surgery, and $Y_i(1)=0$ means that the same patient would not have a heart attack for at least one year if  with surgery. 

The data live in a joint distribution $p(D, A, \mbX, Y(0), Y(1))$ with half of the potential outcomes missing. The missingness is due to the fact that a patient can only be observed under one of the two possible treatments. Thus, only the potential outcome following the observed treatment is observed ($Y_i = Y_i(D_i)$). We introduce ways to estimate the missing potential outcome in \Cref{sec:algorithm}. For now, assume both potential outcomes are handed over to us.

Principal fairness is introduced by Imai and Jiang \citep{imai2023principal}. It works in the potential outcomes framework of causal inference \citep{rubin1974estimating,imbens2015causal}. It assesses the fairness of decisions among individuals who would be similarly affected by the decision. The similarity of individuals is defined based on their response to the decision, captured in the term called principal strata. Principal strata is a causal concept from  principal stratification \citep{frangakis2002principal}. Individuals from the same principal stratum have the same joint potential outcomes $(Y(0), Y(1))$. Let $R_i$ be the principal stratum of patient $i$, $R_i = (Y_i(0), Y_i(1))$. For a binary treatment and a binary outcome, there are four principal strata: $(Y(0) = 0, Y(1) = 0)$ called “both work”, $(Y(0) = 1, Y(1) = 0)$ called “favor treatment”, $(Y(0) = 0, Y(1) = 1)$ called “favor comparator”, and $(Y(0) = 1, Y(1) = 1)$ called “neither works”.

\begin{definition}[Principal fairness] A treatment decision satisfies principal fairness with respect to the outcome of interest $Y_i$ and the sensitive attribute $A_i$ if the decision is independent of the sensitive attribute among patients from the same principal stratum, i.e., $p(D_i \g R_i, A_i) = p(D_i \g R_i)$.
\end{definition}

Principal fairness states that a decision satisfies principal fairness if the decision is independent of the sensitive attribute conditioning on the principal strata, $p(D_i \g R_i, A_i) = p(D_i \g R_i)$. In our example, this means that if men and women have an equal chance of being treated conditioning on their likelihood to benefit (ie., their principal strata), then the decision is fair. Note that this definition allows patients from different principal strata to have a different probability of being treated without violating fairness.  

Based on the definition of principal fairness, the level of violation of principal fairness can be measured by comparing the decision probabilities between the two sensitive groups within a stratum, that is,

\begin{equation}
\begin{aligned}
    \Delta(r) = p(D_i = 1 \g A_i = a, R_i = r)
    - p(D_i = 1 \g A_i = a', R_i = r).
\label{eq:delta}
\end{aligned}
\end{equation}
When $\Delta(r)=0$ for all $r$, principal fairness is satisfied. Otherwise, principal fairness is violated. 

\paragraph{Notes about principal strata}
As \citet{imai2023principal} mentioned, there are some characteristics of principal strata that are important for understanding their role in principal fairness. First, principal strata are \emph{pre-treatment} characteristics of patients. Principal strata are defined based on potential outcomes, not the observed outcome. In other words, potential outcomes are inherent characteristics of a patient. The treatment assignment only changes which potential outcome would be observed, and does not change the value of the potential outcomes.

\paragraph{Notes about principal fairness}
First, principal fairness is a population-level fairness notion. A decision that satisfies population-level fairness may not satisfy fairness at individual level, and vice versa. Second, because principal fairness is based on potential outcomes, which are quantities unobserved by clinicians at the point of decision-making, principal fairness should not be interpreted as a judgment about clinicians' behavior, but rather a retrospective measure of fairness in a system.

\subsection{Causal assumptions}
The key in assessing principal fairness is estimating the conditional probability $p(D \g Y(0), Y(1), A)$. In the original formulation, Imai and Jiang \cite{imai2023principal} assumed that $Y_i(1) \leq Y_i(0)$, an assumption called monotonicity. That is, a patient being treated would always have a lower risk than without the  treatment. Under this assumption, principal fairness, even though a causal quantity, can be identified from observational data \citep{imai2023principal}. Though this assumption may be plausible in the criminal justice setting where Imai and Jiang applied principal fairness to assess the fairness of judge's decisions \citep{imai2021experimental}, monotonicity is unlikely to hold in many situations in healthcare. For example, some treatments' effects could depend on the patient's baseline characteristics (known as heterogeneous treatment effect). In this case, it is unreasonable to assume that one treatment is always better than the other treatment. Instead of assuming monotonicity, we identify an alternative assumption under which principal fairness can be identified and estimated across all principal strata.

\begin{assumption}\label{asp:conditionalIndept}(Conditional independence of potential outcomes). 
$$Y(1) \indpt Y(0)\g \mbX $$
\end{assumption}

\Cref{asp:conditionalIndept} assumes that the two potential outcomes are conditionally independent given the observed covariates. Because we never observe the two potential outcomes jointly, this is an untestable assumption. In the healthcare context, this means that given all the baseline characteristics we know about a patient, further knowing one potential outcome (e.g., survive with treatment) does not provide any additional information about what the other potential outcome would be (e.g, whether the patient would survive or die under no treatment). In addition, we also make the ignorability assumption as in standard causal inference literature.

\begin{assumption} \label{asp:ignore}(Ignorability). 
$$Y(1), Y(0) \indpt D \g \mbX $$
\end{assumption}

\Cref{asp:ignore} assumes that there is no unmeasured confounder for the treatment-outcome relationship. Under ignorability, 
\begin{align*}
    \E{Y(d)} = \EE{X}{\EE{Y}{Y \g \mbX=\mbx, D=d}},
\end{align*} 
or equivalently, for a binary outcome, 
\begin{align*}
    p(Y(d)) = \int p(Y \g x, d)p(x) dx.
\end{align*} 
Under these assumptions, the conditional probability $p(D \g Y(0), Y(1), A)$ in assessing principal fairness can be written as

\begin{align*}
    p(D \g Y(0), Y(1), A) &= \frac{p(D, Y(0), Y(1) \g A)}{p(Y(0), Y(1) \g A)}\\
\end{align*}

where 
\begin{equation}\label{eq:causalToObs}
\begin{aligned}
    p(D, Y(0), Y(1) \g A) &= \int p(D \g X, A)p(Y \g D=0, X, A)p(Y\g D=1, X, A)p(X|A) dx, \\ 
    p(Y(0), Y(1) \g A) &= \int p(Y \g D=0, X, A)p(Y\g D=1, X, A)p(X|A) dx.
\end{aligned}
\end{equation}

In practice, the integral is approximated with Monte Carlo sampling. Notice that the r.h.s. of \Cref{eq:causalToObs} includes only observational distributions that can be estimated from observational data. The proofs are provided in \Cref{appendix:proofs}.

We highlight a choice we make in the ignorability assumption by conditioning on the observed covariates $\mbX$ without conditioning on the sensitive attribute $A$. This choice can be understood from the popular view that variables like race and gender are social constructs rather than biological constructs \citep{fuentes2019aapa}. Take race as an example. The statement that race is a social construct means that race does not have a \emph{direct} causal relationship with the health outcome; In other words, the impact of race on health must be transmitted through mediators (e.g., clinical presentation, social determinants of health). This should not be confused with the relationship between race and treatment decisions: social constructs like race can have a \emph{direct} influence on treatment decisions, that is when the decision maker takes into account race into decision-making. With this perspective of race and gender as social constructs, they are not confounders if all mediators in the causal pathway between race and health outcome are observed, then the ignorability assumption holds by conditioning on all the observed variables only. When there are unobserved variables in the causal pathway, for example, social determinants of health, sensitivity analyses are recommended to assess the robustness of fairness findings. We conducted an empirical sensitivity analysis by comparing the fairness estimates with and without including SDoH in the covariates (see Section \ref{sec:empirical}).

\subsection{Estimation algorithm}\label{sec:algorithm}

Denote the potential outcome model as $\mu^d(x, a) = p(y=1 \g x, a, d)$ for $d \in \{0,1\}$, and denote the treatment probability model as $e(x, a) = p(d=1 \g x, a)$. The algorithm works as follows.

\begin{enumerate}
    \item Inputs: $\cD = \{\mbX_i, A_i, D_i, Y_i\}$
    \item Fit a model on the treatment group to estimate $\mu^1(x, a)$.
    \item Fit a model on the control group to estimate $\mu^0(x, a)$.
    \item Fit a treatment probability model on the entire cohort $e(x, a)$.
    \item Estimate the target conditional probabilities $p(D=1 \g r, a)$ following \Cref{eq:causalToObs}, where $r \in \{(0,0), (0,1), (1,0), (1,1)\}$ and  $a \in \{0,1\}$.
    \item Outputs: $p(D=1 \g r, a), \ \forall r \in \{(0,0), (0,1), (1,0), (1,1)\}, \ a \in \{0,1\}$.
\end{enumerate}

\section{Simulation}\label{sec:simulation}
We simulate a dataset to show that the proposed algorithm can correctly assess whether a decision satisfies principal fairness. Simulation is necessary for evaluation because the ground truth for both potential outcomes is available in a simulation, and never available in any real datasets.

\paragraph{Setup} We simulate the sensitive attribute (e.g. gender) as $A_i \sim \text{Bern}(0.5)$, and pre-treatment covariates as $\mbX_i \sim \cN_{m}(0,1)$, $m=100$. Then, we simulate the potential outcome corresponding to each treatment $d$ for each patient as a function of their pre-treatment covariates $\mbX_i$, that is, $Y_i(d) \sim \text{Bern}(\sigma(f(\mbX_i,d)))$, where $d \in \{0, 1\}$. The function $f(\dot)$ can take any form and we use a linear function without loss of generality. Notice that the potential outcomes do not depend on $A$, which means that no group is inherently healthier or sicker than others given all measured covariates $\mbX$. The principal stratum for each individual $R_i$ is assigned based on each individual's joint potential outcomes, $(Y_i(0), Y_i(1))$. Then, treatment is assigned based on the stratum-specific probability of treatment, $D_i \sim \text{Bern}(p_{r,a})$, where $p_{r,a}$ is the treatment probability for principal stratum $r$ and sensitive attribute $a$. To induce unfairness in treatment allocation in some of the strata, we simulate the treatment such that the probability of being treated is 20\% higher for males than females in the stable stratum ($r=0$), and 20\% lower for males than females in the severe stratum ($r=3$). Details about the simulation setup are in the \Cref{appendix:sim}.


\subsection{Results}

\begin{figure}[htbp!]
    \centering
    \begin{subfigure}[b]{0.7\textwidth}
        \centering
        \includegraphics[width=\textwidth]{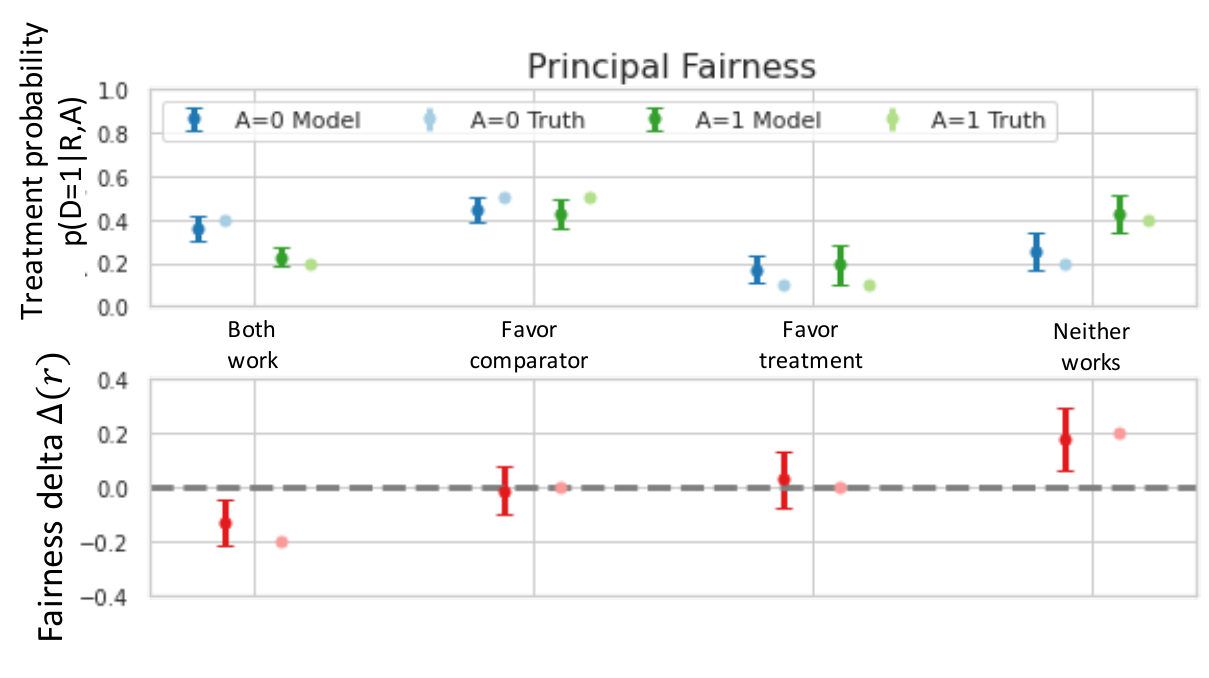}
        \caption{Principal fairness}
        \label{fig:sim_pfdelta}
    \end{subfigure}
    \hfill
    \begin{subfigure}[b]{0.7\textwidth}
        \centering
        \includegraphics[width=\textwidth]{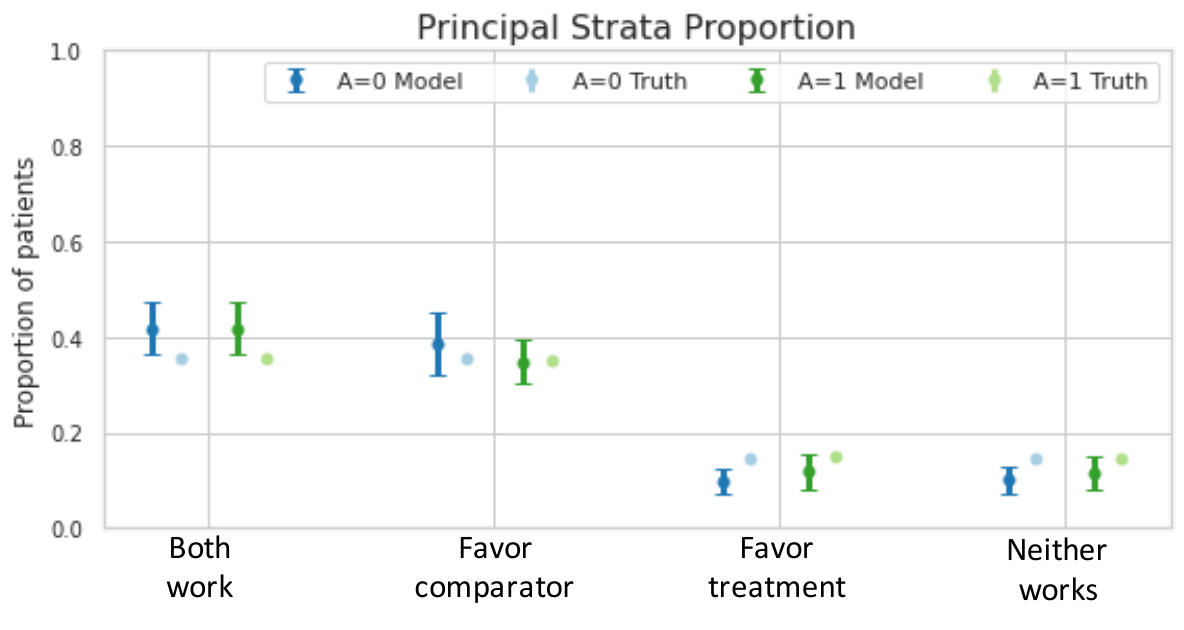}
        \caption{Principal strata proportion}
        \label{fig:sim_pfstrata}       
    \end{subfigure}

    \caption{Principal fairness of decision in simulation. 
    (a): Principal fairness decision probabilities and level of violation metric $\Delta(r)$. (b): Proportion of principal strata. The proposed algorithm can detect unfair decisions and accurately estimate levels of violation.}
    \label{fig:sim}
\end{figure}

The results of the simulation are shown in Figure 1. The proposed algorithm is able to detect the unfair decision and estimate the level of unfairness $\Delta(r)$. The proposed algorithm correctly identified the two strata ("both work" and "neither works") where decisions were made unfairly. Specifically, it was estimated that women are about 20\% less likely to receive the treatment than men in the "both work" group, and 20\% more likely to receive the treatment in the "neither works" group, as shown by the bottom graph of Figure 1(a). In addition, the proposed algorithm correctly estimated the probability of treatment within each principal stratum for men and women (Fig 1(a) top graph), and correctly estimated the proportion of patients within each principal stratum (Figure 1(b)). 

The simulation confirms that the proposed algorithm is able to estimate principal fairness.

\section{Empirical Study} \label{sec:empirical}

We assess the fairness of clinical decisions on revascularization procedures in patients with coronary artery disease (CAD). Heart disease is the leading cause of death for men, women, and people of most racial and ethnic groups in the United States \citep{centersfordiseasecontrolandpreventionnationalcenterforhealthstatistics2022multiple}. Coronary heart disease is the most common type of heart disease, killing 382,820 people in 2020 in the United States--that's 1 in every 10 deaths \citep{centersfordiseasecontrolandpreventionnationalcenterforhealthstatistics2022multiple, tsao2023heart}. Revascularization procedures, including percutaneous coronary intervention (PCI) and coronary artery bypass grafting (CABG), are common clinical procedures for treating CAD. Women, African Americans, and Hispanic populations have been found to have lower odds of receiving revascularization treatments and experience worse outcomes \citep{zea-vera2022racial,gusmano2019disparities, brown2008disparities, li2013racial}. In this study, we apply the proposed algorithm along with other associational fairness metrics to assess sex and racial fairness of CABG relative to PIC in CAD patients using EHR data.

\subsection{Study Design}
\paragraph{Database}
Data for this study come from Columbia University Irving Medical Center-New York Presbyterian Hospital (CUIMC-NYP) EHR database. The database includes over 6 million patient records from the early 1980s to the present day. The CUIMC-NYP EHR database includes numerous data and clinical domains, including visits, conditions, procedures, medications, lab tests, vital signs, and problem lists, among others. The database is formatted according to Observational Medical Outcomes Partnership (OMOP) common data model version 5. OMOP common data model is managed by the Observational Health Data Science and Informatics (OHDSI) open-science community to support the development and execution of observational studies across institutions \citep{hripcsak2015observational}. 

\paragraph{Cohort definition and feature extraction} The coronary artery disease (CAD) cohort consists of two groups, the treatment group (CABG) and the comparator group (PCI). The treatment group includes patients with a diagnosis code of coronary arteriorsclerosis within one year prior to receiving CABG treatment, with the date of CABG treatment as the index date. No prior PCI or CABG treatment was allowed prior to the index date. The comparator group is defined similarly with PCI as the treatment. The primary outcome of interest is acute myocardial infarction (AMI) within one year of treatment initiation. 

Pre-treatment patient features were extracted from EHR and census data. We extracted demographics (race, gender, age on index date), diagnoses, medications, and insurance plans within one year prior to the index date from structured EHR data. We also extracted the area deprivation index (ADI) by linking geocoded addresses in the EHR with census data \cite{jiang2022feasibility}. Patients with missing race or gender were excluded from the study.

The final cohort consists of 21,213 patients, including 6,077 (28.6\%) patients treated with CABG and 15,136 (71.4\%) patients treated with PCI. There were 7,260 (34.2\%) women, and 3,304 patients (15.6\%) whose race variable was recorded as Black or African American. 

\subsection{Principal fairness of treatment allocation}
We apply the principal fairness algorithm to assess the fairness of allocating CABG vs PCI to patients with coronary artery disease. We assess fairness with respect to gender and race. Figure 2(a) presents the fairness assessment with respect to gender. Principal fairness indicates that female patients were less likely to receive CABG than male patients, even if they would benefit equally from the treatment. The biggest gender difference in receiving CABG was observed in the "treatment works" group, where women's probability of receiving CABG was 16.8\% [95\%CI: 13.5\%, 20.0\%] while men's was 24.5\% [95\%CI: 20.6\%, 28.3\%], indicating that women were 7.7\% [95\%CI: 4.2\%, 11.2\%] less likely to be treated with CABG than men. A statistically significant difference was also observed for the "both work" group where women were treated less often with CABG. No unfairness was observed for the rest of the groups. In addition, the principal strata proportion shows that more women were in the "neither works" group and fewer women were in the "both work" group relative to men. The "neither works" group is the largest for both men and women, suggesting that the choice of treatment would not change their health outcome (ie., they would always experience AMI within a year).

Figure 2(b) presents the fairness assessment with respect to race. Principal fairness indicates that Black or African American patients were less likely to receive CABG than non-Black or African American patients, even if they would benefit equally from the treatment. Statistically significant difference in the probability of receiving CABG was observed in "both work", "treatment works", and "neither works" groups, with the largest difference observed in the "neither works" group where Black or African American patients were 8.7\% [95\%CI: 6.6\%, 10.8\%] less likely to be treated with CABG. In addition, the principal strata proportion shows that more Black or African American patients were in the "neither works" group and fewer of them were in the "both work" group. The "neither works" group is the largest among the four, suggesting that the choice of treatment would not change their health outcome (ie., they would always experience AMI within a year).

\begin{figure}[htbp!]
    \centering
    \begin{subfigure}[b]{\textwidth}
    \centering
         \includegraphics[width=\textwidth]{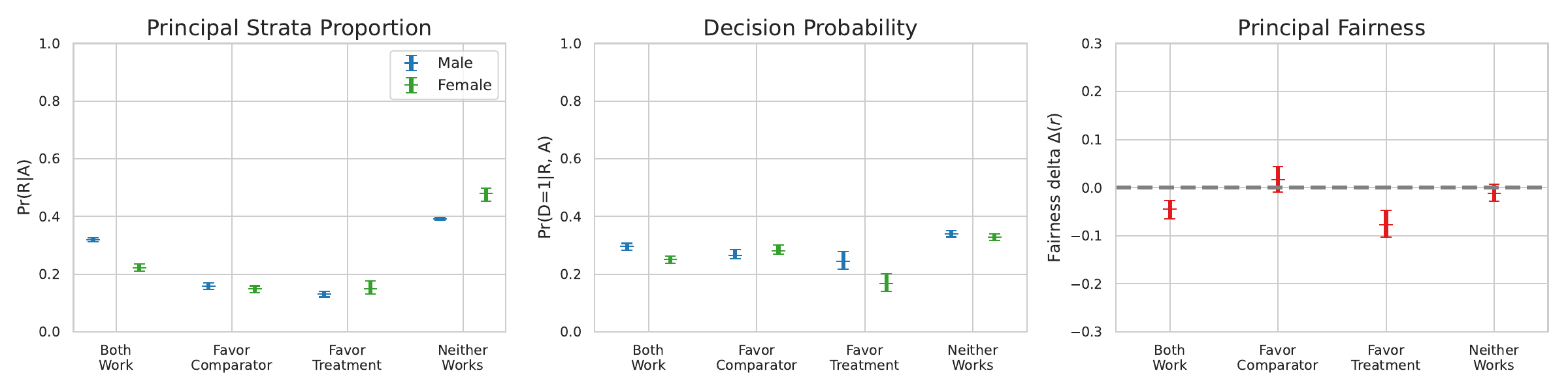}
    	 \caption{Gender principal fairness}
    \end{subfigure}
    \label{fig:principal_fairness_CabgVsPci_gender}
         
    \hfill
         
    \begin{subfigure}[b]{\textwidth}
    \centering
         \includegraphics[width=\textwidth]{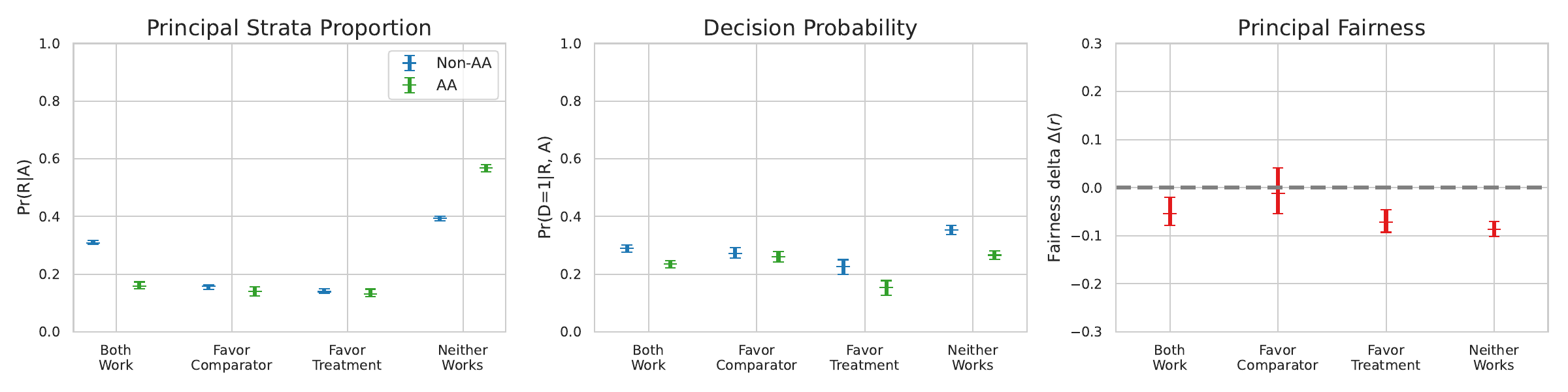}
    	 \caption{Racial principal fairness}        
    \end{subfigure}
\label{fig:principal_fairness_CabgVsPci_race}
\caption{Principal fairness of the allocation of CABG vs PCI. (a) Principal fairness with respect to gender. Negative $\Delta$ indicates that female were treated less frequently with CABG. (b) Principal fairness with respect to race. Negative $\Delta$ indicates that  Black or African American patients were treated less frequently with CABG.} 
\label{fig:principal_fairness_CabgVsPci}
\end{figure}

\subsection{Impact of social determinants of health on principal fairness}

Social determinants of health (SDoH) (e.g., socio-economic status, health insurance coverage) can impact both treatment allocation and health outcomes. SDoH are not well captured in EHR database. When such variables are not available, causal estimates could be  biased. In this study, we empirically evaluate the impact of SDoH on the principal fairness algorithm. We applied the algorithm with and without adjusting for SDoH. Results are shown in \Cref{tab:gender_CabgVsPci_pf} and \Cref{tab:race_CabgVsPci_pf}. Our finding suggests that the fairness estimates were nearly identical between adjusting and not adjusting for SDoH. This does not mean that SDoH is not important for fairness assessment. This empirical similarity suggests that large-scale EHR features are potentially highly correlated with SDoH. Thus, adjusting for large-scale EHR features can implicitly adjust for variables that are not directly included, such as SDoH.

\begin{table}[ht]
\caption{Principal fairness (gender) of revascularization allocation (CABG vs PCI). Results were very similar between with and without adjusting for SDoH. The mean and 95\% bootstrap confidence interval were shown. }
\label{tab:gender_CabgVsPci_pf}
  \resizebox{\textwidth}{!}{
\begin{tabular}{clllll}
\hline
\multicolumn{1}{l}{} &                      & Both Work               & Comparator Works      & Treatment Works         & Neither Works          \\ \hline
\multirow{3}{*}{With SDoH}    & Male & 0.296 (0.279, 0.313) & 0.265 (0.244, 0.286) & 0.245 (0.206, 0.283) & 0.340 (0.323, 0.356) \\
                     & Female               & 0.252 (0.227, 0.277)    & 0.281 (0.250, 0.312)  & 0.168 (0.135, 0.200)    & 0.327 (0.304, 0.351)   \\
                     & Fairness $\Delta(r)$ & -0.044 (-0.067, -0.021) & 0.016 (-0.015, 0.048) & -0.077 (-0.112, -0.042) & -0.013 (-0.037, 0.012) \\ \hline
\multirow{3}{*}{Without SDoH} & Male & 0.297 (0.282, 0.313) & 0.270 (0.247, 0.292) & 0.238 (0.192, 0.284) & 0.343 (0.318, 0.367) \\
                     & Female               & 0.251 (0.228, 0.273)    & 0.284 (0.244, 0.324)  & 0.163 (0.122, 0.204)    & 0.332 (0.311, 0.354)   \\
                     & Fairness $\Delta(r)$ & -0.047 (-0.069, -0.024) & 0.014 (-0.012, 0.041) & -0.076 (-0.108, -0.043) & -0.01 (-0.031, 0.011)  \\ \hline
\end{tabular}
}
\end{table}

\begin{table}[ht]
\caption{Principal fairness (race) of revascularization allocation (CABG vs PCI). Results were very similar between with and without adjusting for SDoH. The mean and 95\% bootstrap confidence interval were shown.}
\label{tab:race_CabgVsPci_pf}
  \resizebox{\textwidth}{!}{
\begin{tabular}{clllll}
\hline
\multicolumn{1}{l}{} &                      & Both Work               & Comparator Works       & Treatment Works         & Neither Works           \\ \hline
\multirow{3}{*}{With SDoH}    & Non-AA & 0.289 (0.272, 0.306) & 0.272 (0.248, 0.296) & 0.227 (0.192, 0.262) & 0.353 (0.334, 0.372) \\
                     & AA                   & 0.235 (0.199, 0.270)    & 0.260 (0.215, 0.305)   & 0.154 (0.133, 0.175)    & 0.266 (0.256, 0.276)    \\
                     & Fairness $\Delta(r)$ & -0.054 (-0.093, -0.016) & -0.012 (-0.067, 0.043) & -0.072 (-0.102, -0.043) & -0.087 (-0.108, -0.066) \\ \hline
\multirow{3}{*}{Without SDoH} & Non-AA & 0.290 (0.274, 0.306) & 0.274 (0.247, 0.302) & 0.221 (0.177, 0.264) & 0.357 (0.334, 0.380) \\
                     & AA                   & 0.230 (0.194, 0.265)    & 0.273 (0.227, 0.319)   & 0.149 (0.119, 0.178)    & 0.269 (0.248, 0.290)    \\
                     & Fairness $\Delta(r)$ & -0.061 (-0.101, -0.02)  & -0.002 (-0.047, 0.044) & -0.072 (-0.095, -0.048) & -0.088 (-0.107, -0.069) \\ \hline
\end{tabular}
}
\end{table}

\section{Discussion} \label{sec:discussion}

In this study, we develop an algorithm to explore the potential of a causal fairness notion called principal fairness in assessing the fairness of treatment decisions. The proposed approach assesses the fairness of treatment allocation by stratifying patients based on their treatment responses while adjusting for a large set of baseline patient characteristics. 

Many fairness metrics have been proposed in the discrimination discovery literature. Statistical parity \citep{dwork2012fairness}, equality of opportunity, mistreatment parity, and predictive equality \citep{hardt2016equality, zafar2016fairness, corbett-davies2017algorithmic} are the most frequently reviewed associational metrics. Recently, a growing number of fairness notions are based on causality, reflecting the widely accepted idea that causal reasoning is essential for addressing the problem of fairness. By viewing discrimination as the presence of an unfair causal effect of the sensitive attribute on the decision, \citet{qureshi2020causal} presents a method for causal discrimination discovery that adjusts for confounding using propensity score analysis. Some causal fairness takes a step further to distinguish direct and indirect discrimination based on path-specific effects. \citet{zhang2019causal, zhang2016causal} leverage path-specific effects to discover and remove direct and indirect discrimination from observational data. \citet{nabi2018fair, zhang2018mitigating, wang2019blessingsa} developed various methods to quantify direct and indirect discrimination. \citet{kilbertus2017avoiding} proposed discrimination criteria to qualitatively determine the existence of indirect discrimination. \citet{huan2020fairness} proposed to assess fairness by quantifying the difference in effort to achieve the same outcome. \citet{kusner2017counterfactual} introduced an individual-level causal fairness criterion called counterfactual fairness, which
states that a decision is fair toward an individual if it is the same as the decision that would have been taken in a counterfactual world where the sensitive attribute was different. Counterfactual fairness and principal fairness consider different variables as the intervention. Counterfactual fairness intervenes on the sensitive attribute directly, while principal fairness assesses fairness based on potential outcomes under a different medical treatment, and then uses this causal quantity to further assess fairness. It is intuitively more approachable to estimate the potential outcome with respect to medical treatment than with respect to a sensitive attribute. Another difference is that principal fairness is population-level fairness, while counterfactual fairness is individual-level, but can be population-level with some modifications. The two levels of fairness do not imply each other \citep{imai2021experimental}. 

Leveraging established fairness metrics commonly used in predictive models, \citet{sun2022assessing} proposed a set of best practices to assess the fairness of phenotype definitions and related algorithmic fairness metrics to commonly used epidemiological cohort description metrics. \citet{pfohl2019counterfactual} developed an augmented counterfactual fairness criterion that extends the group fairness criteria of equalized odds for clinical risk prediction. The importance of fair machine learning for healthcare is emphasized in several perspectives and commentaries along with proposed guidelines \citep{chen2021ethical, gichoya2021equity, ghassemi2020review}, but the gap between machine learning, fairness, and healthcare is still huge and needs to be filled to advance health equity.

There are several limitations to this approach. First, the proposed algorithm for assessing principal fairness relies on assumptions for causal identification. Our approach requires Assumption 1, conditional independence of the potential outcomes conditioned on the available covariates. One alternative is assuming monotonicity \citep{imai2023principal} such that treatment is never harmful or the target treatment is always better than the comparator, but this is rarely true in medicine. Another approach is modeling the two potential outcomes with Bayesian multi-task learning \citep{alaa2017bayesian}, but it, too, carries assumptions about prior distributions, and it requires a more complex computational approach. Another assumption required by our approach is ignorability, which assumes all confounders are observed. This is a standard yet untestable assumption in causal studies. In studies using EHR data, adjusting for all available concepts without confounder selection has been shown to be empirically effective in handling unmeasured confounding bias \citep{zhang2022adjusting}. Negative controls have also been used to empirically assess residual bias \citep{lipsitch2010negative, schuemie2018empirical}. In more general settings, sensitivity analysis based on simulations is recommended to assess robustness \citep{hernan2020causal}.

Second, the causes of disparity remain to be understood. Exploring the reasons for the disparity in treatment allocation could yield opportunities to improve health equity. To do so, researchers should keep in mind that treatment decision for CAD is often a team decision, including cardiologists, cardiac surgeons, patients and their families \citep{lawton20222021}, and influenced by system-level (e.g., hospital capacity)\citep{li2013racial} and society-level factors. EHR data alone as they are now is unlikely to be sufficient for such studies due to the lack of more granular information capturing the decision-making process, and natural language processing from clinical notes plus additional system-level data may fill in the missing data required for such a study.

Third, the proposed algorithm focuses on assessing treatment disparities, while health care is a dynamic process, factors that precede treatment decision-making, such as access to care, diagnosis disparities, and testing bias can potentially have an impact on the treatment decision. Future work should look into how to extend principal fairness to account for bias in other stages of care delivery using sequential models.

Last but not least, this work is subject to all limitations regarding the use of EHR for observational research \citep{hripcsak2013nextgeneration}. In particular, the not-at-random missingness of race in half of the patient population in the EHR database can affect the fairness, validity, and generalizability of the method and the results. 

\section{Conclusion} \label{sec:conclusion}
We develop a causal fairness algorithm for assessing treatment allocation with electronic health records. We illustrate our approach by assessing the fairness of coronary artery revascularization allocation in patients with coronary artery disease. We uncover disparities in revascularization treatment allocation based on gender and race, with women and Black or African American patients being less likely to receive coronary artery bypass grafting, highlighting the importance of addressing fairness concerns in clinical practice. Furthermore, we demonstrate that social determinants of health, variables that are often unavailable in EHR databases and are potential unmeasured confounders, do not significantly impact the estimation of treatment responses when a large set of clinical features from EHRs are used in estimating treatment responses, shedding light on the intricate correlation between EHR features and social determinants of health. 

\section*{Author contributions}
All authors contributed to the conceptualisation of the proposed method, though Linying Zhang had a leading role in this work. All authors contributed to draft writing, editing and review.

\section*{Funding}
This work was supported by National Institutes of Health (NIH) R01LM006910.

\section*{Declaration of Competing Interest}
The authors declare that they have no known competing financial interests or personal relationships that could have appeared to influence the work reported in this paper.

\bibliographystyle{elsarticle-num-names} 
\bibliography{biblio}

\clearpage
\appendix
\section{Proofs of principal fairness}\label{appendix:proofs}

Proofs for the estimation of principal fairness. Recall that we make the following assumptions.
\begin{assumption} (Unconfoundedness). 
$$Y(1), Y(0) \indpt D \g \mbX $$
\end{assumption}

\begin{assumption}(Conditional independence of potential outcomes). 
$$Y(1) \indpt Y(0)\g X $$
\end{assumption}

Our target estimand is

\begin{align*}
    p(D \g Y(0), Y(1), A) &= \frac{p(D, Y(0), Y(1) \g A)}{p(Y(0), Y(1) \g A)}\\
\end{align*}

The denominator is
\begin{align*}
    p(Y(0), Y(1) \g A) &= \int p(Y(0), Y(1) \g X, A)p(X|A) dx\\
    &= \int p(Y(0) \g X, A)p(Y(1)\g X, A)p(X|A) dx\\
    &= \int p(Y \g D=0, X, A)p(Y\g D=1, X, A)p(X|A) dx\\
    &\approx \frac{1}{n_a} \sum_{i:A_i=a}p(Y \g D=0, X, A)p(Y \g D=1, X, A)
\end{align*}

The first equality holds because of law of total probability. The second holds because of Assumption 1. The third equality holds because of Assumption 2.

The numerator is
\begin{align*}
    p(D, Y(0), Y(1) \g A) &= \int p(D, Y(0), Y(1) \g X, A)p(X|A) dx\\
    &= \int p(D \g X, A)p(Y(0), Y(1) \g X, A)p(X|A) dx\\
    &= \int p(D \g X, A)p(Y(0) \g X, A)p(Y(1) \g X, A)p(X|A) dx\\
    &\approx \frac{1}{n_a} \sum_{i:A_i=a}p(D \g X, A)p(Y \g D=0, X, A)p(Y \g D=1, X, A)
\end{align*}

The first equality holds because of law of total probability. The second holds because of Assumption 2. The third equality holds because of Assumption 1.

\clearpage

\section{Simulation Details}\label{appendix:sim}
We simulate a data set to demonstrate the effectiveness of the algorithm in assessing the fairness of decisions. The benefit of a simulated dataset is that we have access to the ground truth (i.e., both potential outcomes for all individuals), which is not available in a real clinical setting. We simulate the data as follows:
\begin{enumerate}
    \item Simulate a binary sensitive attribute as $A \sim \text{Bern}_n(0.5)$.
    \item Simulate covariates as $\mbX \sim \cN_{n\times m}(0,1)$, where $n=5,000$ is the number of patients and $m=100$ is the number of covariates.
    \item Simulate potential outcomes as 
    \begin{align*}
        Y_i(0) \sim \text{Bern}(\text{sigmoid}(\mbx_i^\top \theta_{y_0} + \theta_d 0))\\
        Y_i(1) \sim \text{Bern}(\text{sigmoid}(\mbx_i^\top \theta_{y_1} + \theta_d 1))
    \end{align*}
    where $\theta_{y_0}, \theta_{y_1} \sim \cN_{m}(0,1)$. The effect size of the treatment $\theta_d = -1$.
    \item Assign patients to principal strata.
    \begin{equation*}
  R_i =
    \begin{cases}
      0 \text{(stable),} & \text{if } Y_i(0),Y_i(1) = 0,0  \\
      1 \text{(treatable),} & \text{if } Y_i(0), Y_i(1) = 1,0\\
      2 \text{(better-wo),} & \text{if } Y_i(0),Y_i(1) = 0,1\\
      3 \text{(severe),} & \text{if } Y_i(0),Y_i(1) = 1,1.
    \end{cases}   
\label{eq:strata}
\end{equation*} 
    \item Simulate decision $D_i$ conditioning on principal strata and the sensitive attribute as $D_i \g R_i, A_i \sim \text{Bern}(p_{h,a}))$ where $\Delta(r) = p_{h,a} -p_{h,a'} = 0$ for $h=1,2$, and $\Delta(r) = p_{h,1} - p_{h,0} = -0.2$ for $h=0$ and $\Delta(r) = p_{h,1} - p_{h,0} = 0.2$ for $h=3$. That is, the decision is unfair in two of the four principal strata, and specifically, the decision favors individuals with $A=0$ in the stable stratum but favors individuals with $A=1$ in the severe stratum.
\end{enumerate}

\clearpage

\section{Association-based fairness metrics}
We show the results from three popular association-based fairness metrics: statistical parity, accuracy, and calibration. All three fairness metrics detect differences in the delivery of revascularization across gender and race. We will interpret the results for gender as an example. The interpretation of race is analogous. 

When assessing the fairness of revascularization between men and women, statistical parity indicates that male patients are more likely to receive CABG than female patients. Calibration indicates that the health outcome (acute myocardial infarction) happens at a higher rate for male patients than for female patients in both CABG and PCI groups. Accuracy indicates male patients are more likely to receive CABG than female patients in both outcome groups. The differences shown by these metrics do not allow conclusions to be made regarding the fairness of treatment assignment, because whether there is any health difference at the baseline between men and women is not known.

\begin{table}[htbp!]
  \resizebox{\textwidth}{!}{\begin{tabular}{llllll}
  \hline
                            & \multicolumn{2}{c}{A = Male} & \multicolumn{2}{c}{A = Female} &          \\
                            \hline
\textbf{Statistical parity} & Mean   & 95\% CI             & Mean    & 95\% CI              & p-value  \\
p(D=CABG $\g$A)                 & 0.296  & {[}0.288, 0.303{]}  & 0.269   & {[}0.259, 0.279{]}   & 3.17E-05 \\
\hline
\textbf{Accuracy}           & Mean   & 95\% CI             & Mean    & 95\% CI              & p-value  \\
p(D=CABG $\g$A, Y=AMI)          & 0.326  & {[}0.315, 0.337{]}  & 0.304   & {[}0.290, 0.317{]}   & 0.01     \\
p(D=CABG $\g$A, Y=No AMI)       & 0.263  & {[}0.253, 0.274{]}  & 0.209   & {[}0.194, 0.225{]}   & 1.30E-08 \\
\hline 
\textbf{Calibration}        & Mean   & 95\% CI             & Mean    & 95\% CI              & p-value  \\
p(Y=AMI $\g$A, D=CABG)          & 0.568  & {[}0.552, 0.583{]}  & 0.712   & {[}0.692, 0.733{]}   & 3.86E-29 \\
p(Y=AMI $\g$A, D=PCI)           & 0.492  & {[}0.482, 0.502{]}  & 0.601   & {[}0.587, 0.614{]}   & 9.08E-38 \\
\hline
\end{tabular}
}
\caption{Fairness of revascularization allocation between men and women based on association-based fairness metrics.}
\label{tab:asso_gender}
\end{table}

\begin{table}[htbp!]
  \resizebox{\textwidth}{!}{\begin{tabular}{llllll}
  \hline
                            & \multicolumn{2}{c}{A=Non-African American} & \multicolumn{2}{c}{A=African American} &          \\
                            \hline
\textbf{Statistical parity} & Mean          & 95\% CI                    & Mean        & 95\% CI                  & p-value  \\
p(D=CABG $\g$A)                 & 0.296         & {[}0.290, 0.303{]}         & 0.234       & {[}0.219, 0.248{]}       & 1.52E-14 \\
\hline
\textbf{Accuracy}           & Mean          & 95\% CI                    & Mean        & 95\% CI                  & p-value  \\
p(D=CABG $\g$A, Y=AMI)          & 0.332         & {[}0.323, 0.342{]}         & 0.256       & {[}0.238, 0.273{]}       & 9.96E-14 \\
p(D=CABG $\g$A, Y=No AMI)       & 0.256         & {[}0.246, 0.265{]}         & 0.183       & {[}0.159, 0.207{]}       & 4.02E-08 \\
\hline
\textbf{Calibration}        & Mean          & 95\% CI                    & Mean        & 95\% CI                  & p-value  \\
p(Y=AMI $\g$A, D=CABG)          & 0.592         & {[}0.579, 0.606{]}         & 0.763       & {[}0.733, 0.793{]}       & 2.40E-23 \\
p(Y=AMI $\g$A, D=PCI)           & 0.501         & {[}0.492, 0.509{]}         & 0.678       & {[}0.660, 0.696{]}       & 8.90E-64 \\
\hline
\end{tabular}
}
    \caption{Fairness of revascularization allocation between African American and non-African American patients based on association-based fairness metrics.}
    \label{tab:asso_race}
\end{table}

\clearpage

\section{Principal fairness algorithm with random forest architecture}
We also implemented a \textit{random forest} model for estimating principal fairness. Here are the results when we applied the random forest model to the CAD cohort to quantify the fairness of treatment allocation. The results were similar to that from the regression model presented in the paper, and the conclusions regarding fairness stay the same.

\begin{table}[htbp!]
\caption{Principal fairness (gender) of revascularization allocation (CABG vs PCI). Results were very similar between with and without adjusting for SDoH. The model is \textit{random forest}. The mean and 95\% bootstrap confidence interval were shown.}
\label{tab:gender_CabgVsPci_pf_randomforest}
  \resizebox{5.5in}{!}{
\begin{tabular}{clllll}
\hline
\multicolumn{1}{l}{} &                      & Both Work               & Comparator Works     & Treatment Works         & Neither Works          \\ \hline
\multirow{3}{*}{With SDoH}    & Male & 0.276 (0.271, 0.282) & 0.297 (0.286, 0.307) & 0.189 (0.183, 0.195) & 0.356 (0.344, 0.367) \\
                     & Female               & 0.239 (0.230, 0.249)    & 0.351 (0.336, 0.365) & 0.128 (0.120, 0.136)    & 0.339 (0.326, 0.351)   \\
                     & Fairness $\Delta(r)$ & -0.037 (-0.045, -0.029) & 0.054 (0.033, 0.075) & -0.061 (-0.066, -0.057) & -0.017 (-0.035, 0.001) \\ \hline
\multirow{3}{*}{Without SDoH} & Male & 0.276 (0.271, 0.281) & 0.304 (0.296, 0.313) & 0.192 (0.184, 0.200) & 0.352 (0.338, 0.365) \\
                     & Female               & 0.239 (0.229, 0.249)    & 0.344 (0.330, 0.359) & 0.130 (0.123, 0.138)    & 0.339 (0.325, 0.353)   \\
                     & Fairness $\Delta(r)$ & -0.037 (-0.042, -0.032) & 0.04 (0.019, 0.061)  & -0.062 (-0.072, -0.051) & -0.012 (-0.027, 0.002) \\ \hline
\end{tabular}
}
\end{table}

\begin{table}[htbp!]
\caption{Principal fairness (race) of revascularization allocation (CABG vs PCI). Results were very similar between with and without adjusting for SDoH. The model is \textit{random forest}. The mean and 95\% bootstrap confidence interval were shown.}
\label{tab:race_CabgVsPci_rf_randomforest}
  \resizebox{5.5in}{!}{
\begin{tabular}{clllll}
\hline
\multicolumn{1}{l}{} &                      & Both Work               & Comparator Works        & Treatment Works         & Neither Works           \\ \hline
\multirow{3}{*}{With SDoH}    & Non-AA & 0.272 (0.266, 0.278) & 0.316 (0.308, 0.325) & 0.180 (0.173, 0.188) & 0.362 (0.354, 0.371) \\
                     & AA                   & 0.215 (0.204, 0.225)    & 0.301 (0.287, 0.315)    & 0.106 (0.099, 0.114)    & 0.294 (0.276, 0.312)    \\
                     & Fairness $\Delta(r)$ & -0.058 (-0.072, -0.044) & -0.016 (-0.031, -0.0)   & -0.074 (-0.086, -0.062) & -0.068 (-0.084, -0.053) \\ \hline
\multirow{3}{*}{Without SDoH} & Non-AA & 0.271 (0.265, 0.277) & 0.320 (0.311, 0.329) & 0.181 (0.172, 0.189) & 0.362 (0.353, 0.372) \\
                     & AA                   & 0.214 (0.208, 0.220)    & 0.303 (0.291, 0.316)    & 0.114 (0.106, 0.123)    & 0.286 (0.269, 0.302)    \\
                     & Fairness $\Delta(r)$ & -0.057 (-0.067, -0.047) & -0.017 (-0.033, -0.001) & -0.066 (-0.079, -0.053) & -0.076 (-0.09, -0.063)  \\ \hline
\end{tabular}
}
\end{table}

\clearpage

\section{Model performance}

We evaluated the goodness-of-fit by area under the curve (AUC) of Receiver Operator Characteristic Curve and Precision Recall Curve, and also F1-score. The algorithm is l2-regularized logistic regression. 

\begin{table}[!htbp]
\caption{Potential outcome model performance (Gender fairness + SDOH)}
\label{tab:performance_gender_fairness_sdoh}
  \resizebox{5.5in}{!}{
\begin{tabular}{lllllll}
 \hline
           & ROC\_AUC &                & PR\_AUC &                & F1\_score &                \\  
           & mean     & 95\%CI         & mean    & 95\%CI         & mean      & 95\%CI         \\ \hline
$f_0(x,a)$ & 0.893    & (0.887, 0.898) & 0.853   & (0.847, 0.859) & 0.905     & (0.899, 0.910) \\
$f_1(x,a)$ & 0.921    & (0.913, 0.929) & 0.915   & (0.907, 0.923) & 0.945     & (0.939, 0.951) \\  \hline
\end{tabular}
}
\end{table}

\begin{table}[!htbp]
\caption{Potential outcome model performance (Gender fairness - SDOH)}
\label{tab:performance_gender_fairness_nosdoh}
  \resizebox{5.5in}{!}{
\begin{tabular}{lllllll}
 \hline
           & ROC\_AUC &                & PR\_AUC &                & F1\_score &                \\
           & mean     & 95\%CI         & mean    & 95\%CI         & mean      & 95\%CI         \\  \hline
$f_0(x,a)$ & 0.888    & (0.881, 0.895) & 0.845   & (0.838, 0.852) & 0.902     & (0.897, 0.908) \\
$f_1(x,a)$ & 0.918    & (0.912, 0.925) & 0.912   & (0.906, 0.919) & 0.944     & (0.939, 0.949) \\  \hline
\end{tabular}
}
\end{table}

\begin{table}[!htbp]
\caption{Potential outcome model performance (Racial fairness + SDOH)}
\label{tab:performance_racial_fairness_sdoh}
  \resizebox{5.5in}{!}{
\begin{tabular}{lllllll}
 \hline
           & ROC\_AUC &                & PR\_AUC &                & F1\_score &                \\
           & mean     & 95\%CI         & mean    & 95\%CI         & mean      & 95\%CI         \\  \hline
$f_0(x,a)$ & 0.893    & (0.887, 0.898) & 0.853   & (0.847, 0.859) & 0.905     & (0.899, 0.910) \\
$f_1(x,a)$ & 0.921    & (0.913, 0.929) & 0.915   & (0.907, 0.923) & 0.945     & (0.939, 0.951) \\  \hline
\end{tabular}
}
\end{table}

\begin{table}[!htbp]
\caption{Potential outcome model performance (Racial fairness - SDOH)}
\label{tab:performance_racial_fairness_nosdoh}
  \resizebox{5.5in}{!}{
\begin{tabular}{lllllll}
 \hline
           & ROC\_AUC &                & PR\_AUC &                & F1\_score &                \\
           & mean     & 95\%CI         & mean    & 95\%CI         & mean      & 95\%CI         \\  \hline
$f_0(x,a)$ & 0.888    & (0.881, 0.895) & 0.845   & (0.838, 0.852) & 0.902     & (0.897, 0.908) \\
$f_1(x,a)$ & 0.918    & (0.912, 0.925) & 0.912   & (0.906, 0.919) & 0.944     & (0.939, 0.949) \\  \hline
\end{tabular}
}
\end{table}
\onecolumn

\end{document}